\documentclass[10pt, conference]{IEEEtran}
\IEEEoverridecommandlockouts

\usepackage[pdftex]{graphicx}
    \graphicspath{{/}{fig/}{../fig}}
    \DeclareGraphicsExtensions{.pdf,.jpeg,.png}

\usepackage{cite}
\usepackage[cmex10]{amsmath}
\usepackage{array}
\usepackage{textcomp}
\usepackage{xcolor}
\usepackage{multirow}
\usepackage{amssymb}
\usepackage{mathtools}
\usepackage{float}
\usepackage[vlined, ruled, shortend]{algorithm2e}
\usepackage{booktabs}
\usepackage{upgreek}
\usepackage{fancyref}

\usepackage{pgfplots}

\pgfplotsset{compat=1.7}

\usepackage{flushend}

\newlength\figureheight
\newlength\figurewidth
\title{
    Multi Sensor Fusion for Navigation and Mapping in Autonomous Vehicles: Accurate Localization in Urban Environments
}

\author{
    \IEEEauthorblockN{
        Li Qingqing\textsuperscript{1,2},
        Jorge Peña Queralta\textsuperscript{2},
        Tuan Nguyen Gia\textsuperscript{2},
        Zhuo Zou\textsuperscript{1} and
        Tomi Westerlund\textsuperscript{2}
    }\\[-4pt]
    \IEEEauthorblockA{
        \textsuperscript{1} School of Information Science and Technology, Fudan Universtiy, China \\
        \textsuperscript{2} Department of Future Technologies, University of Turku, Finland \\
        Emails: \{qingqingli16, zhuo\}@fudan.edu.cn, \\ \{jopequ, tunggi, tovewe\}@utu.fi \\
    }
}

\begin{document}

\maketitle
\thispagestyle{empty}
\pagestyle{empty}

\begin{abstract}

The combination of data from multiple sensors, also known as sensor fusion or data fusion, is a key aspect in the design of autonomous robots. In particular, algorithms able to accommodate sensor fusion techniques enable increased accuracy, and are more resilient against the malfunction of individual sensors. The development of algorithms for autonomous navigation, mapping and localization have seen big advancements over the past two decades. Nonetheless, challenges remain in developing robust solutions for accurate localization in dense urban environments, where the so called last-mile delivery occurs. In these scenarios, local motion estimation is combined with the matching of real-time data with a detailed pre-built map. In this paper, we utilize data gathered with an autonomous delivery robot to compare different sensor fusion techniques and evaluate which are the algorithms providing the highest accuracy depending on the environment. The techniques we analyze and propose in this paper utilize 3D lidar data, inertial data, GNSS data and wheel encoder readings. We show how lidar scan matching combined with other sensor data can be used to increase the accuracy of the robot localization and, in consequence, its navigation. Moreover, we propose a strategy to reduce the impact on navigation performance when a change in the environment renders map data invalid or part of the available map is corrupted.


\end{abstract}


\IEEEpeerreviewmaketitle
\section{Introduction}

Accurate mapping and localization is the cornerstone of self-driving cars. In open roads or highways, lane-following can partly replace localization while still allowing for autonomous operation~\cite{visual_lidar_map}. Nonetheless, in dense urban environments accurate localization is a paramount aspect of a robot's autonomous operation~\cite{robust_ndt}. Smaller pathways and more dynamic environments pose significant technical challenges~\cite{levinson2007map}. In addition, the robot mission often adds accuracy requirements, such as in autonomous post delivery~\cite{autonomous_delivery_patent}. Multiple sensors can be utilized to facilitate autonomous navigation and operation. Among those, visual data provides more semantic and qualitative information~\cite{se2005vision, garcia2015vision
}, lidar measurements are more accurate and are able to accurately describe objects from a geometric point of view~\cite{premebida2016high}.

\begin{figure}
    \centering
    \includegraphics[width=0.49\textwidth]{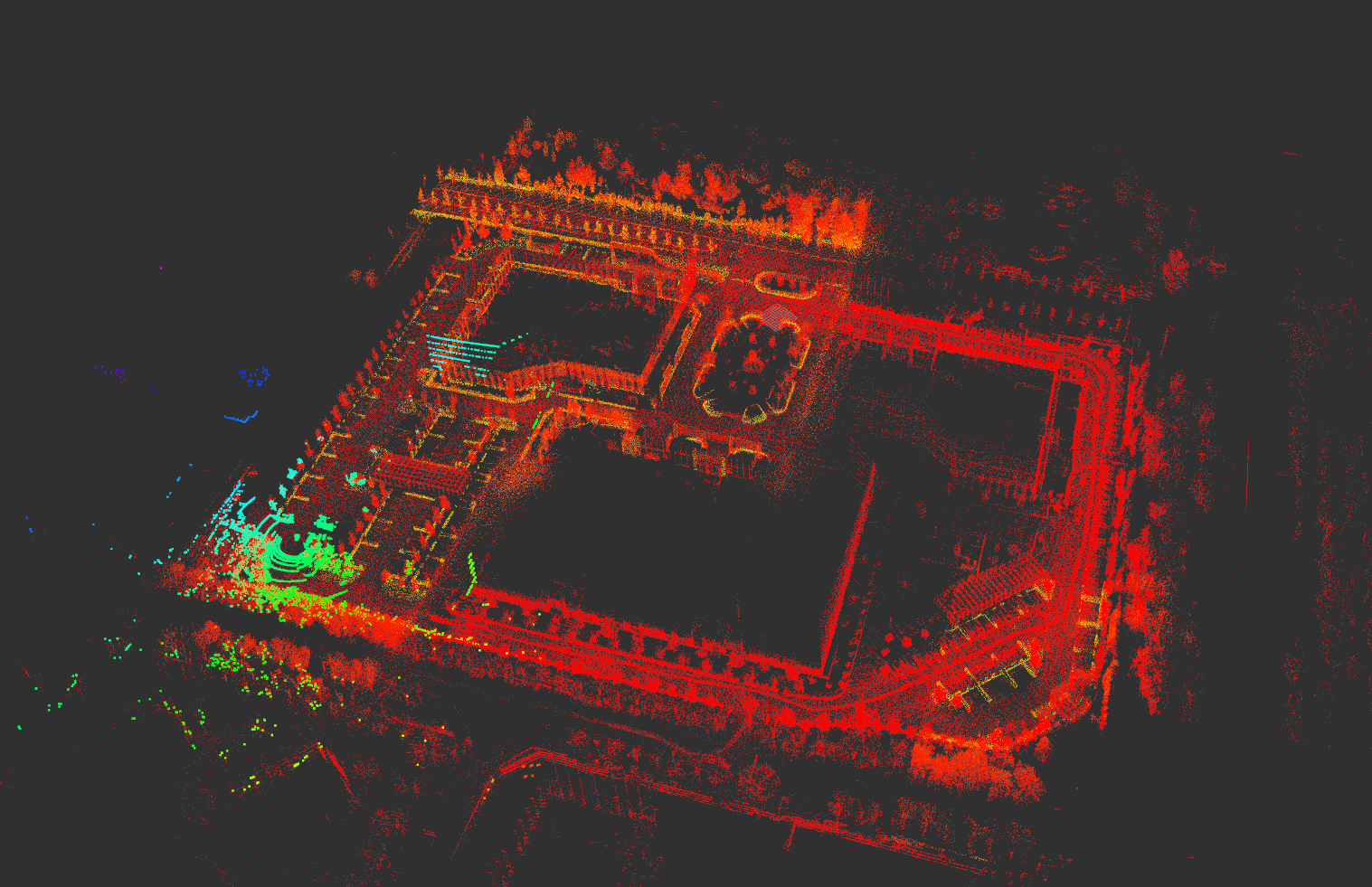}
    \caption{Illustration of the matching process between the pre-acquired map (red) and current lidar scan (green-blue).}
    \label{fig:localization_example}
\end{figure}

The past decade has seen a boost in the development of autonomous vehicles for civilian use. Google started the development of its self-driving technology for cars in 2009~\cite{john2010google}, and since then a myriad of industry leaders~\cite{bojarski2016end, harris2015documents}, start-ups~\cite{greenblatt2016self}, and academic researchers~\cite{kato2015open} have joined the race in the technology sector, a race to make everything autonomous. In any mobile robot or vehicle, SLAM algorithms are an essential and crucial aspect of autonomous navigation~\cite{lidar_road_detection, robust_ndt}.

Autonomous robots or self-driving cars will potentially disrupt the logistics industry worldwide~\cite{hofmann2017industry}. Autonomous trucks or autonomous cargo vessels are already in advanced stages or development and might be seen in operation within the next five or ten years~\cite{flamig2016autonomous}. However, both technological and legal challenges remain within the so called "last-mile" delivery~\cite{boysen2018scheduling}. Last-mile refers to the last step in the delivery of goods from a local logistics or supply center to the clients' door. In this paper, we utilize data gathered using a small delivery robot from Jingdong, one of the top two e-commerce platform and B2C online retailer in China.

The development of simultaneous localization and mapping (SLAM) algorithms has seen a rapid evolution over the past two decades~\cite{zhang2014loam, zhang2015visual}. In SLAM algorithms, information from a wide range of sensors can be used to map the environment and localize the vehicle or robot in real time. These include inertial measurement units, monocular or binocular cameras, GNSS sensors, lidars, ultrasonic sensors or radars, or wheel encoders~\cite{sensor_fusion_survey}. Detailed 3D maps in the form of point clouds can be generated, for instance, from 3D lidars or with stereo vision~\cite{premebida2016high}. We focus on the study and comparison of different localization methods for a small delivery robot in dense urban environments. In these scenarios, an existing map of the 
operating environment is obtained in advance and either pre-loaded or accessible by the autonomous robot. The map is used in order to obtain more accurate localization by matching each
scan with a certain area of the map in real-time~\cite{ndt_differential, robust_ndt}.

The local motion of a robot or vehicle can be estimated directly by integrating data from inertial measurement units, including accelerometers, gyroscopes and compasses. Alternatively, different odometry methods can be applied based on non-inertial sensors. Visual odometry algorithms utilize feature extraction and tracking, while lidar-based odometry uses mostly geometric information~\cite{zhang2014loam, zhang2015visual}. Inertial measurement units can be easily combined with wheel encoders. Differential GNSS measurements also provide accurate local motion estimation~\cite{rizos2012precise}. Global localization can be estimated either with GNSS data, or by comparing sensor data with predefined maps or information gathered a priori. For instance, different methods exist to match a lidar scan with section of a 3D point cloud that defines a map of the operational area~\cite{ndt_differential, robust_ndt}. Over the past decades, researchers from both industry and academia have been exploring the utilization of these methods and their combinations to obtain accurate mapping and localization. More concretely, scholars often refer to the combination of different sensor data as sensor fusion or data fusion. In this paper, we compare different techniques and provide arguments on the best sensor fusion techniques for a small delivery robot for last-mile delivery.

The algorithms, analysis and results presented in this paper were mostly developed during the JD Digital (JDD) Globalization Challenge Competition in "Multi-sensor fusion localization for autonomous driving". The challenge was a global competition, with 4 classification divisions depending on the geographical location of the team. Our team ranked first in the US division semi-final and classified for the 24h global final in Beijing, China, where the 4 semi-final winners competed for the first prize. The available sensor data during the competition was GNSS and gyroscope data, wheel odometry and the output from a 16-channel 3D lidar. A map of the area was given as a 3D point cloud. Multiple datasets exist to test and benchmark different localization algorithms. However, the most accurate algorithms are obtained through fine tuning and parameters specific for the dataset, with different parameters being potentially necessary to achieve the optimal accuracy in a different sensor or environment setup~\cite{kitti_benchmarking}. Therefore, in this paper we have utilized the data provided as part of the JDD competition as it was gathered from the sensors on-board the vehicle in order to compare a variety of methods. This ensures that our algorithms can be implemented on the same robot without a significant impact to performance.

The main contributions of this work are the following: (i) to analyze and compare different approaches for vehicle localization estimation, and the definition of a sensor fusion approach for accurate localization in urban environments; and (ii) the introduction of a strategy for rebuilding an area of a local map when data is corrupted or the environment has undergone significant modifications.

The remainder of this paper is organized as follows: Section II introduces related work in localization and odometry techniques based on data available from the on-board sensors. Section III describes the dataset used in the paper, delves into the possibilities of sensor fusion approaches based on the available information, and provides a strategy to operate in areas where the available map is either outdated or corrupted. Then, section IV shows experimental results on the localization accuracy for four different sensor fusion approaches. Finally, Section V concludes the work and outlines directions for future work.

\section{Related Work}

Autonomous navigation through 3D mapping with lidars has been an increasingly popular technology over the last decade, as lidars can provide high accuracy range measurements when compared to other sensors. Zhang \textit{et al.} proposed a method for lidar odometry and mapping. The authors approached the odometry problem by extracting lidar point cloud features from each sweep. Then,  the transformation between two consecutive sweeps can be estimated. In this setup, the lidar is utilized as an odometry sensor~\cite{zhang2014loam}. For 3D mapping, the most important problem is accurate position estimation, and 3d lidar data for odometry still can produce accumulated error after long distance walk. However, the accumulated error is much lower when compared to camera-based odometry (especially in unfavourable light conditions) or integration of inertial data from accelerometers/gyroscopes. State-of-the art solutions combine lidar and monochrome camera sensors as visual-lidar odometry to improve the performance of ego-motion estimation~\cite{zhang2015visual}.

In SLAM algorithms, localization and mapping are done concurrently in real time. However, in order to achieve localization with centimeter accuracy in an urban environment, map matching techniques have emerged over the past few years. Currently, one of the most widely used approaches for 3D point cloud matching is Normal Distributions Transform (NDT) matching~\cite{biber2003normal}. Introduced by Bibel \textit{et al.}, NDT has the advantages of no requiring explicit assignments of relationships between features or subsets of points, and the analytic formulation of the algorithms. The former aspect increases the robustness of the algorithm, while the latter reduces the computational cost and accuracy of the implementation.

\begin{figure*}
    \centering
    \includegraphics[width=\textwidth]{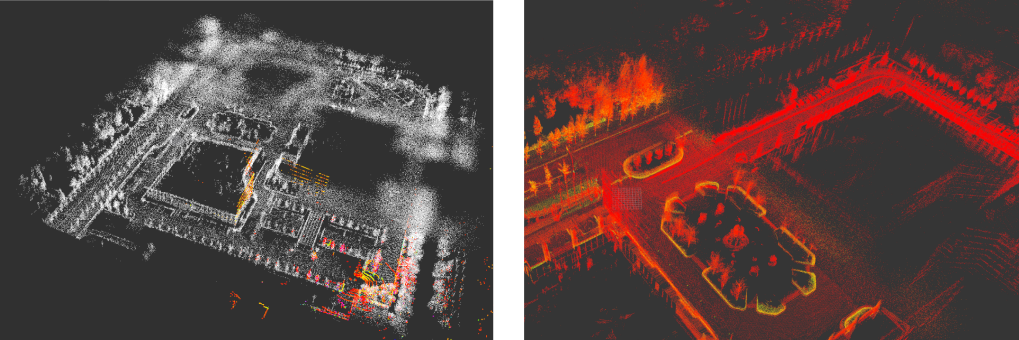} \\
    \caption{On the left, a map with noise added in some areas to simulate corrupted data. On the right, one of the two corrupted sections of the the map has been restored using GNSS, IMU and lidar odometry.}
    \label{fig:fixing_the_map}
\end{figure*}

Multiple improvements of the NDT algorithm have been proposed. Gonzalez Prieto \textit{et al.} presented DENDT, an algorithm for 3D-NDT scan matching with Differential Evolution~\cite{ndt_differential}. The authors utilize a differential evolution (DE) algorithm in order to improve the optimization process for finding the solution of the NDT method. Akai \textit{et al.} presented a robust localization method that uses intensity readings from the lidar data in order to detect road markings and use them for matching consecutive scans~\cite{robust_ndt}. While the method is able to provide very high accuracy in some environments, the existence of road markers significantly impacts the performance and therefore this method cannot be used in all situations. Wen \textit{et al.} recently analyzed the performance of different NDT-based SLAM algorithms in multiple scenarios in Hong Kong~\cite{wen2018performance}. A valuable conclusion from this work is that the best performance was achieved in areas with more sparse point clouds and nominal traffic conditions, while the performance decreased in dense urban areas. In this work, we study how can we combine different sensor information in order to improve localization performance in an urban area.

\section{Localization and Mapping}

In this section, we describe the dataset that we have utilized and the different localization approaches. We also introduce a strategy for situations where the existing map data is corrupted, outdated, or part of the data is missing.

\subsection{Dataset}

The data utilized in this paper was provided as part of the JD Discovery Global Digitalization Challenge from December 2018 to January 2019. The data was gathered using JD's autonomous last-mile delivery robot. The data includes: (1) GNSS directional and positional data referenced in the World Geodetic System (WGS84) format; (2) lidar data as a 3D point cloud; (3) raw accelerometer and gyroscope data; and (4) wheel speed meter. Ground truth data is provided as well. The output of the 3D lidar is given at 10~Hz, IMU data is acquired at 100~Hz and GNSS data is updated at 5~Hz. In addition, a map of the objective operation area is given. The map represented as a point cloud is shown in red in Fig.~\ref{fig:localization_example}. The dataset contains sensor data recorded in an 800-second long closed loop movement.

In order to both read and process data, ROS has been utilized. ROS (Robot Operating system) is an open source operating system for robots, which provides a publish-subscribe communication framework that allows for rapid development of distributed robotic systems~\cite{quigley2009ros}. ROS provides algorithm reuse support for robot research and development, as well as abstraction of data models for easier integration of different modules. PCL (Point Cloud Library) is a cross-platform open source C++ library, which implements common algorithms and data structures of point clouds~\cite{pointcloudlibrary}. It can realize point cloud acquisition, filtering, segmentation, registration, retrieval, feature extraction, recognition, tracking, surface reconstruction, visualization and so on. If OpenCV is the crystallization of 2D visual data acquisition and processing, PCL has the same position in the 3D geometrical data domain.

\subsection{Localization methods}

Based on the available sensor data, we have utilized five different approaches to estimate the vehicle's localization. In each approach, we use a different combination of sensors and describe how the robot position is calculated based on their data.

\vspace{8pt}
\noindent\emph{GNSS-based localization}

One of the most traditional methods for outdoor robot localization is to use a global navigation satellite system. In this case, data from multiple satellite constellations was available and used for increase accuracy. GNSS data error are mostly caused by the atmospheric conditions and multi-path interference. The effect of the environment in a larger scale and the atmospheric conditions can be minimized using differential GNSS readings, and assuming that the real-time error is equivalent to the error obtained in a near known location with which the system is synchronized. However, in this work we have not relied on differential GNSS.

\vspace{8pt}
\noindent\emph{GNSS+IMU localization}

We can easily combine GNSS data with inertial data, including both accelerometer and accuracy. As differential GNSS has not been implemented in this case, instead, the results labelled as "IMU" utilize the IMU readings for local motion estimation, and the GNSS reading for an initial global estimation and estimations when the robot movement is almost zero for a prolonged period of time.

\vspace{10pt}
\noindent\emph{Lidar odometry (LOAM)}

Zhang \textit{et al.} introduced lidar odometry as an alternative to the more classical visual odometry techniques \cite{zhang2014loam}. As with many odometry approaches, features are extracted from data and compared within consecutive frames, or scans in the case of a lidar. Features extracted from lidar data are usually based on geometrical aspects. These include corners and surfaces, for instance. Because lidars are able to provide high accuracy distance measurements even for objects far away from the sensors, lidar-based odometry is able to provide higher accuracy than visual-based odometry in open space situations with clearly differentiated objects. An implementation based on Zhang's algorithm has been used in this case.

\vspace{10pt}
\noindent\emph{NDT-based localization (NDT+)}

The NDT algorithm is a kind of registration algorithm that uses the existing high-precision point cloud map and real-time 3D lidar point cloud data to achieve high-precision localization.

NDT algorithm does not directly compare the distance between points in point clouds map and points in lidar point clouds. First, the NDT algorithm will transform the point cloud map into the normal distribution in three-dimensional.  

If a variable $X$ is normal distribution $X \sim (\mu,\delta)$, then it can be described as:

\begin{equation}
    f(x)= \frac{1}{\delta \sqrt{2\pi}} e^{ \frac{-(x-\mu)^2}{2\delta^2}}
\end{equation}

where $\mu$ is the mean of the variable distribution and $\delta^2$ is the variance. For a multivariate normal distribution, its probability density function can be expressed as:

\begin{equation}
    f(\Vec{x})= \frac{1}{({2\pi})^{\frac{D}{2}} \sqrt{\left | \sum \right |}} e^{ -(\Vec{x}-\Vec{\mu})^{T} \sum^{-1}(\Vec{x}-\Vec{\mu})} 
\end{equation}

Where $\Vec{x}$ represents the mean vector and $\sum$ represents the covariance matrix.

The first step of the NDT algorithm is to divide the point cloud into a 3D grid coordinate. For each cell, the probability distribution function(PDF) is calculated based on the points distribution density in the grid.

\begin{equation}
    \Vec{\mu} = \frac{1}{m} \sum_{k=1}^{m} \Vec{y_k}
\end{equation}
\begin{equation}
    \Upsigma = \frac{1}{m} \sum_{k=1}{m}(\Vec{y_k}-\Vec{\mu})(\Vec{y_k}-\Vec{\mu})^T
\end{equation}

Where $\Vec{y_k} = 1,2,3, m$ denotes all lidar points in a grid. Then the PDF can be expressed as:

\begin{equation}
    f(\Vec{x})= \frac{1}{({2\pi})^{\frac{3}{2}} \sqrt{\left| \Upsigma \right| }} e^{ -(\Vec{x}-\Vec{\mu})^{T} \sum^{-1}(\Vec{x}-\Vec{\mu})}
\end{equation}

We use the normal distribution to represent the discrete points of each grid. Each probability density function can be considered as an approximation of a local surface. It not only describes the location of the surface in space but also contains information about the direction and smoothness of the surface.

After calculated the PDF of each grid, then our goal is to find the best transformation. The lidar point cloud set is $X= \Vec{x_1},\Vec{x_2},...,\Vec{x_n} $, and the parameter of transformation is $\Vec{p}$. The spatial transformation function $T(\Vec{p},\Vec{x_k})$ represents using transformation $\Vec{p}$ to move point $\Vec{x_k}$, combined with the previous calculated state density function(the PDF of each grid), so the best transformation $\Vec{p}$ should be the transformation of maximum likelihood function:
    $$ Likelihood: \theta = \prod_{k=1}^{n}f(T(\Vec{p}, \Vec{x_k}))$$
And the maximum likelihood is equivalent to minimum negative logarithmic likelihood:
$$ -\log{\theta} =  -\sum_{k=1}^{n}\log{f(T(\Vec{p}, \Vec{x_k}))}$$

The task now becomes to minimize the negative logarithmic likelihood by using an optimization algorithm to adjust the transformation parameter $\Vec{p}$. we can use the Newton method to optimize the parameters. 

The main problem of the NDT approach is its stability when used standalone. As indicated by the authors of previous works, NDT alone has the disadvantage of being unstable depending on the scenario \cite{wen2018performance}. Therefore, we utilize GNSS data for setting the initial position as well as resetting the NDT localization method when a sudden change in position or orientation estimation is detected. In the results, we refer to this method as NDT+, and a close implementation to the one provided in existing NDT method has been used \cite{wen2018performance}.

\vspace{10pt}
\noindent\emph{NDT+IMU localization (NDT++)}

The final localization utilized in our experiments consisted on integrating the IMU data into the NDT+ method described above that uses lidar and GNSS data. With this approach, we have been able to eradicate the instabilities of the NDT+ method and increase its accuracy. 

The algorithm workflow is as follows: first, on system start-up or reset, GNSS data is used in order to obtain an initial estimation of the robot's location. This estimation can be utilized in order to reduce the area of the map in which the NDT matching will be looked for. Second, when the robot starts moving, an unscented Kalman filter that uses IMU data as input serves as an estimation between lidar scans. The Kalman filter output is then feeded to the NDT algorithm for scan matching with the predefined map. The GNSS data is still used to avoid instabilities, even though we have not detected any in the dataset utilized in this paper.

The NDT+ and NDT++ approaches have an additional benefit over the lidar odometry method. In autonomous robots moving in an urban environment, it is essential to react on time to obstacles and to have localization information as frequently as possible. A lidar-only approach has the disadvantage of receiving sensor updates at only 10Hz in this case. With IMU readings having a refresh rate of 100Hz, the IMU can be utilized to obtain local movement estimation between lidar scan matches using the NDT approach. This minimizes the possibilities of instabilities in the NDT algorithm as the matching possibilities are reduced and the goal of the algorithm partly shifts from coarse localization to increasing the accuracy of IMU-based movement estimation.

\subsection{Corrupted Map Reconstruction}

In an urban scenario, it is impractical to propose a localization method that has a high dependency on the existence of an accurate map of the operational area without a strategy for operating in case the map data is corrupted or outdated. In Fig.~\ref{fig:fixing_the_map}, we show the map of the operation area (on the left, in black and white), with two areas where the data has either been removed completely or noise has been added to render the NDT algorithm unusable. When the robot approaches these areas, we are able to detect them by monitoring the difference between the NDT localization and GNSS and IMU positioning. When part of the map cannot be matched with current scans, we utilize lidar odometry and mapping in order to rebuild the corrupted or missing data. The result of this process is shown on the right side of Fig.~\ref{fig:fixing_the_map}, where one of the corrupted map areas has been restored in real-time while the robot was travelling through it using lidar odometry and mapping. Even though it is not visible in the image, there is a relatively small mismatch in the map in the area where the robot emerges again into a mapped environment.

\begin{figure}[t]
    \centering
    \setlength\figureheight{0.35\textwidth}
    \setlength\figurewidth{0.49\textwidth}
    \scriptsize{\input{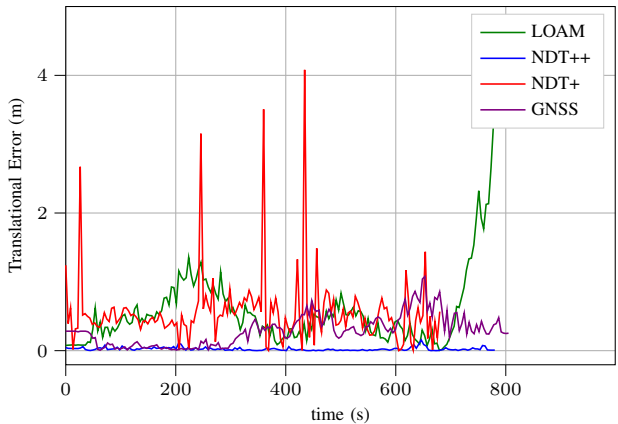}} \\
    \caption{Translational errors of the proposed approaches over time.}
    \label{fig:errors}
\end{figure}

\begin{figure}[t]
    \centering
    \setlength\figureheight{0.35\textwidth}
    \setlength\figurewidth{0.49\textwidth}
    \scriptsize{\input{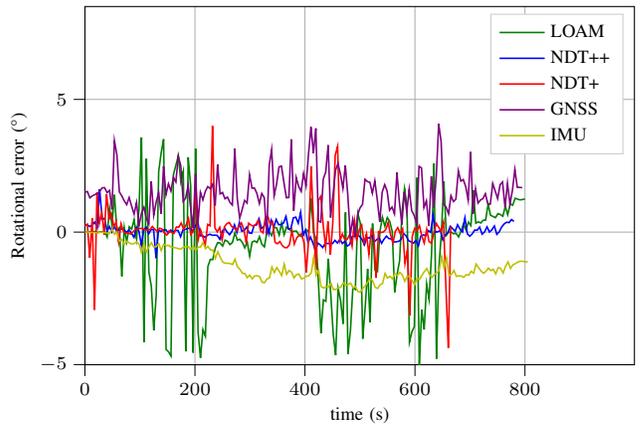}} \\
    \caption{Rotational errors of the proposed approaches over time.}
    \label{fig:yaw_errors}
\end{figure}

\begin{table}[t]
    \centering
    \caption{Localization error mean and stadard deviation}
    \begin{tabular}{@{}lcccccc@{}}
        \toprule
        \textbf{} & 
        \textbf{$\mu_{{rot.}}$} & 
        \textbf{$\sigma_{rot}$} &
        \textbf{$\mu_{{x}}$} & 
        \textbf{$\sigma_x$} &
        \textbf{$\mu_{{y}}$} & 
        \textbf{$\sigma_y$} \\
        \midrule
        \textbf{GNSS}   & 1.52 & 0.79  & -0.46  & 0.22  & $10^{-4}$  & 0.18 \\
        \textbf{IMU}    & -1.24 & 0.62  & N/A  & N/A  & N/A  & N/A \\
        \textbf{LOAM}   & -0.50 & 1.88  & 0.40  & 0.49  & 0.10  & 0.49 \\
        \textbf{NDT+}   & 0.03 & 0.87  & -0.02  & 1.51  & \textbf{-0.05}  & 1.13 \\
        \textbf{NDT++}  & \textbf{$10^{-3}$} & \textbf{0.31}  & \textbf{-0.01}  & 0.10  & \textbf{-0.05 } & \textbf{0.10 }\\
        \bottomrule
    \end{tabular}
    \label{tab:results}
\end{table}

\section{Experiment and Results}

We have applied the five approaches proposed to the given dataset. The results showing the translational and rotational localization and orientation error are shown in Fig.~\ref{fig:errors} and Fig.~\ref{fig:yaw_errors}, respectively. In these figures, the NDT++ method shows a stable and very small error though time, both in position and rotation estimation. The NDT+ without taking into account inertial data shows a larger error but, more importantly, shows several instabilities that are corrected from the GNSS data.

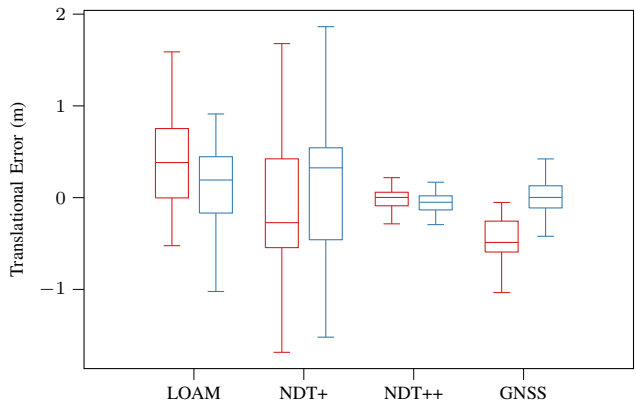
\begin{figure}[t]
    \centering
    \setlength\figureheight{0.35\textwidth}
    \setlength\figurewidth{0.49\textwidth}
    \scriptsize{
\begin{tikzpicture}

\definecolor{color0}{rgb}{0.843137254901961,0.0980392156862745,0.109803921568627}
\definecolor{color1}{rgb}{0.172549019607843,0.482352941176471,0.713725490196078}

\begin{axis}[
height=\figureheight,
legend cell align={left},
legend style={draw=white!80.0!black},
tick align=outside,
tick pos=left,
width=\figurewidth,
x grid style={white!69.01960784313725!black},
xmin=-2, xmax=8,
xtick style={color=black},
xtick={0,2,4,6},
xticklabels={LOAM,NDT+,NDT++,GNSS},
y grid style={white!69.01960784313725!black},
ylabel={Translational Error (m)},
ymin=-1.86317725921683, ymax=2.04077551845403,
ytick style={color=black}
]
\addplot [color0, forget plot]
table {%
-0.7 -0.00309944821609331
-0.1 -0.00309944821609331
-0.1 0.752841396044012
-0.7 0.752841396044012
-0.7 -0.00309944821609331
};
\addplot [color0, forget plot]
table {%
-0.4 -0.00309944821609331
-0.4 -0.523440888178385
};
\addplot [color0, forget plot]
table {%
-0.4 0.752841396044012
-0.4 1.5888406642963
};
\addplot [color0, forget plot]
table {%
-0.55 -0.523440888178385
-0.25 -0.523440888178385
};
\addplot [color0, forget plot]
table {%
-0.55 1.5888406642963
-0.25 1.5888406642963
};
\addplot [color0, forget plot]
table {%
1.3 -0.544836737593765
1.9 -0.544836737593765
1.9 0.422762047819608
1.3 0.422762047819608
1.3 -0.544836737593765
};
\addplot [color0, forget plot]
table {%
1.6 -0.544836737593765
1.6 -1.68572486023179
};
\addplot [color0, forget plot]
table {%
1.6 0.422762047819608
1.6 1.67876681400199
};
\addplot [color0, forget plot]
table {%
1.45 -1.68572486023179
1.75 -1.68572486023179
};
\addplot [color0, forget plot]
table {%
1.45 1.67876681400199
1.75 1.67876681400199
};
\addplot [color0, forget plot]
table {%
3.3 -0.0886290566698698
3.9 -0.0886290566698698
3.9 0.0587289747826567
3.3 0.0587289747826567
3.3 -0.0886290566698698
};
\addplot [color0, forget plot]
table {%
3.6 -0.0886290566698698
3.6 -0.285471794387902
};
\addplot [color0, forget plot]
table {%
3.6 0.0587289747826567
3.6 0.217663526758386
};
\addplot [color0, forget plot]
table {%
3.45 -0.285471794387902
3.75 -0.285471794387902
};
\addplot [color0, forget plot]
table {%
3.45 0.217663526758386
3.75 0.217663526758386
};
\addplot [color0, forget plot]
table {%
5.3 -0.592609747289604
5.9 -0.592609747289604
5.9 -0.256074527605023
5.3 -0.256074527605023
5.3 -0.592609747289604
};
\addplot [color0, forget plot]
table {%
5.6 -0.592609747289604
5.6 -1.03476551324131
};
\addplot [color0, forget plot]
table {%
5.6 -0.256074527605023
5.6 -0.0523811531369915
};
\addplot [color0, forget plot]
table {%
5.45 -1.03476551324131
5.75 -1.03476551324131
};
\addplot [color0, forget plot]
table {%
5.45 -0.0523811531369915
5.75 -0.0523811531369915
};
\addplot [color1, forget plot]
table {%
0.1 -0.168073443544264
0.7 -0.168073443544264
0.7 0.445863408464163
0.1 0.445863408464163
0.1 -0.168073443544264
};
\addplot [color1, forget plot]
table {%
0.4 -0.168073443544264
0.4 -1.02281626975378
};
\addplot [color1, forget plot]
table {%
0.4 0.445863408464163
0.4 0.91201429789416
};
\addplot [color1, forget plot]
table {%
0.25 -1.02281626975378
0.55 -1.02281626975378
};
\addplot [color1, forget plot]
table {%
0.25 0.91201429789416
0.55 0.91201429789416
};
\addplot [color1, forget plot]
table {%
2.1 -0.4587984620594
2.7 -0.4587984620594
2.7 0.543191336560987
2.1 0.543191336560987
2.1 -0.4587984620594
};
\addplot [color1, forget plot]
table {%
2.4 -0.4587984620594
2.4 -1.5206388887018
};
\addplot [color1, forget plot]
table {%
2.4 0.543191336560987
2.4 1.86332311946899
};
\addplot [color1, forget plot]
table {%
2.25 -1.5206388887018
2.55 -1.5206388887018
};
\addplot [color1, forget plot]
table {%
2.25 1.86332311946899
2.55 1.86332311946899
};
\addplot [color1, forget plot]
table {%
4.1 -0.13377180500702
4.7 -0.13377180500702
4.7 0.0195443425910475
4.1 0.0195443425910475
4.1 -0.13377180500702
};
\addplot [color1, forget plot]
table {%
4.4 -0.13377180500702
4.4 -0.294185519577127
};
\addplot [color1, forget plot]
table {%
4.4 0.0195443425910475
4.4 0.168144579446306
};
\addplot [color1, forget plot]
table {%
4.25 -0.294185519577127
4.55 -0.294185519577127
};
\addplot [color1, forget plot]
table {%
4.25 0.168144579446306
4.55 0.168144579446306
};
\addplot [color1, forget plot]
table {%
6.1 -0.112252034247
6.7 -0.112252034247
6.7 0.129196142777005
6.1 0.129196142777005
6.1 -0.112252034247
};
\addplot [color1, forget plot]
table {%
6.4 -0.112252034247
6.4 -0.4211594285443
};
\addplot [color1, forget plot]
table {%
6.4 0.129196142777005
6.4 0.42221041116899
};
\addplot [color1, forget plot]
table {%
6.25 -0.4211594285443
6.55 -0.4211594285443
};
\addplot [color1, forget plot]
table {%
6.25 0.42221041116899
6.55 0.42221041116899
};
\addplot [color0, forget plot]
table {%
-0.7 0.382459935585296
-0.1 0.382459935585296
};
\addplot [color0, forget plot]
table {%
1.3 -0.272722372377984
1.9 -0.272722372377984
};
\addplot [color0, forget plot]
table {%
3.3 0.00230953645868837
3.9 0.00230953645868837
};
\addplot [color0, forget plot]
table {%
5.3 -0.488381883979002
5.9 -0.488381883979002
};
\addplot [color1, forget plot]
table {%
0.1 0.192040383920812
0.7 0.192040383920812
};
\addplot [color1, forget plot]
table {%
2.1 0.324736052192804
2.7 0.324736052192804
};
\addplot [color1, forget plot]
table {%
4.1 -0.0502598860435626
4.7 -0.0502598860435626
};
\addplot [color1, forget plot]
table {%
6.1 0.00202515721318974
6.7 0.00202515721318974
};
\end{axis}

\end{tikzpicture}} \\
    \caption{Boxplots of translational errors for the different approaches: x-error (red) and y-error (blue).}
    \label{fig:boxplot_errors}
\end{figure}

\begin{figure}[t]
    \centering
    \setlength\figureheight{0.35\textwidth}
    \setlength\figurewidth{0.49\textwidth}
    \scriptsize{
\begin{tikzpicture}

\definecolor{color0}{rgb}{0.843137254901961,0.0980392156862745,0.109803921568627}

\begin{axis}[
height=\figureheight,
legend cell align={left},
legend style={draw=white!80.0!black},
tick align=outside,
tick pos=left,
width=\figurewidth,
x grid style={white!69.01960784313725!black},
xmin=-2, xmax=10,
xtick style={color=black},
xtick={0,2,4,6,8},
xticklabels={LOAM,NDT+,NDT++,GNSS,IMU},
y grid style={white!69.01960784313725!black},
ylabel={Rotational Error (\textdegree)},
ymin=-4.51685810142521, ymax=3.51310872393692,
ytick style={color=black}
]
\addplot [color0, forget plot]
table {%
-0.6 -1.40390621271525
0.6 -1.40390621271525
0.6 0.455702838348271
-0.6 0.455702838348271
-0.6 -1.40390621271525
};
\addplot [color0, forget plot]
table {%
0 -1.40390621271525
0 -4.1518596093633
};
\addplot [color0, forget plot]
table {%
0 0.455702838348271
0 3.14440012350951
};
\addplot [color0, forget plot]
table {%
-0.3 -4.1518596093633
0.3 -4.1518596093633
};
\addplot [color0, forget plot]
table {%
-0.3 3.14440012350951
0.3 3.14440012350951
};
\addplot [color0, forget plot]
table {%
1.4 -0.287321066415949
2.6 -0.287321066415949
2.6 0.261104818826308
1.4 0.261104818826308
1.4 -0.287321066415949
};
\addplot [color0, forget plot]
table {%
2 -0.287321066415949
2 -0.971395113097998
};
\addplot [color0, forget plot]
table {%
2 0.261104818826308
2 0.729589855957599
};
\addplot [color0, forget plot]
table {%
1.7 -0.971395113097998
2.3 -0.971395113097998
};
\addplot [color0, forget plot]
table {%
1.7 0.729589855957599
2.3 0.729589855957599
};
\addplot [color0, forget plot]
table {%
3.4 -0.178878593884576
4.6 -0.178878593884576
4.6 0.184226316712095
3.4 0.184226316712095
3.4 -0.178878593884576
};
\addplot [color0, forget plot]
table {%
4 -0.178878593884576
4 -0.707477654955014
};
\addplot [color0, forget plot]
table {%
4 0.184226316712095
4 0.636242071165015
};
\addplot [color0, forget plot]
table {%
3.7 -0.707477654955014
4.3 -0.707477654955014
};
\addplot [color0, forget plot]
table {%
3.7 0.636242071165015
4.3 0.636242071165015
};
\addplot [color0, forget plot]
table {%
5.4 1.044960848242
6.6 1.044960848242
6.6 1.88696368263695
5.4 1.88696368263695
5.4 1.044960848242
};
\addplot [color0, forget plot]
table {%
6 1.044960848242
6 -0.130227734374998
};
\addplot [color0, forget plot]
table {%
6 1.88696368263695
6 3.148110231875
};
\addplot [color0, forget plot]
table {%
5.7 -0.130227734374998
6.3 -0.130227734374998
};
\addplot [color0, forget plot]
table {%
5.7 3.148110231875
6.3 3.148110231875
};
\addplot [color0, forget plot]
table {%
7.4 -1.6711663508461
8.6 -1.6711663508461
8.6 -0.638811352152899
7.4 -0.638811352152899
7.4 -1.6711663508461
};
\addplot [color0, forget plot]
table {%
8 -1.6711663508461
8 -2.26738514386601
};
\addplot [color0, forget plot]
table {%
8 -0.638811352152899
8 0.105453426623995
};
\addplot [color0, forget plot]
table {%
7.7 -2.26738514386601
8.3 -2.26738514386601
};
\addplot [color0, forget plot]
table {%
7.7 0.105453426623995
8.3 0.105453426623995
};
\addplot [color0, forget plot]
table {%
-0.6 -0.0141371125498999
0.6 -0.0141371125498999
};
\addplot [color0, forget plot]
table {%
1.4 0.0542248624585042
2.6 0.0542248624585042
};
\addplot [color0, forget plot]
table {%
3.4 0.00798475978139734
4.6 0.00798475978139734
};
\addplot [color0, forget plot]
table {%
5.4 1.45709574589802
6.6 1.45709574589802
};
\addplot [color0, forget plot]
table {%
7.4 -1.44523454692103
8.6 -1.44523454692103
};
\end{axis}

\end{tikzpicture}} \\
    \caption{Boxplots of rotational errors for the different approaches in degrees.}
    \label{fig:boxplot_yaw}
\end{figure}
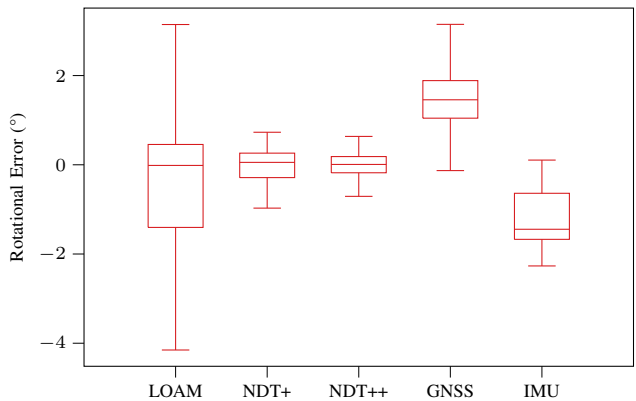

In order to be able to compare in more detail the different methods, and to see whether there exist some background error or drifting, we show the variability of the localization error though two sets of boxplots in Fig.~\ref{fig:boxplot_errors} and Fig.~\ref{fig:boxplot_yaw}. The specific values are also listed in Table \ref{tab:results}. We have omitted the location errors of standalone IMU motion estimation as the error is significantly higher than the proposed approaches. However, inertial data is still valuable for local movement estimation and for orientation estimation.

\begin{figure}
    \centering
    \setlength\figureheight{0.35\textwidth}
    \setlength\figurewidth{0.49\textwidth}
    \scriptsize{\input{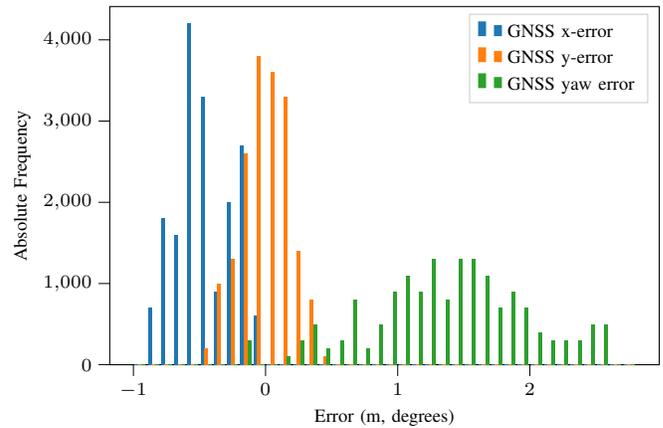}}
    \caption{Distribution of GNSS errors.}
    \label{fig:histogram}
\end{figure}

The translational errors are shown in Fig.~\ref{fig:boxplot_errors}, where the red boxes show the error in the $x$ coordinate, and the blue boxes refer to the $y$ coordinate error. We have separated the error because in some cases the error mean differs between them. That is the case of the GNSS data, which due to atmospheric or environmental conditions shows a steady negative drift in the $x$ direction. If consistent through large periods of time, it can be assumed that it is due to a sensing error in the device itself, or to environment conditions such a specific multi-path occurring in the operating area. Therefore, this value can be utilized to decrease the sensing error in real time during operation. In order to have a deeper understanding of the distribution of the GNSS error, we show the histogram of the three errors (two translational, one rotational) in Fig.~\ref{fig:histogram}. Only the error in the $y$ direction has a mean of 0, while the distribution of the $x$ error is symmetrical and narrow. Therefore, it is possible to fix the drift while keeping the same variance for both components of the translational error. In the case of the rotational error, it is more complex to correct even though the distribution is still symmetric.

From Fig.~\ref{fig:boxplot_errors} and Fig.~\ref{fig:boxplot_yaw} we can see that the most stable methods are NDT++ and GNSS, with the NDT+ method providing accurate results for rotation estimation. However, the NDT+ is highly unstable for position estimation, with the highest variance of all presented approaches. In position estimation, all approaches have a relatively small error after 800 seconds of movement, except for the LOAM method, which error drifts away from 0 towards the end of the available data set. Similarly, the IMU constantly drifts in terms of orientation estimation but it provides a more stable measurement than LOAM, HDL+ or GNSS.

In summary, lidar scan matching with a 3D map provides the highest accuracy for localization, both in terms of position and orientation. Nonetheless, it is essential to take into account other sensor data in order to implement a more robust approach that is less prone to instabilities and depends less on the operational environment. GNSS and inertial data are essential for increasing the localization accuracy but also for minimizing the possibilities of unexpected behaviour in the algorithm.

\section{Conclusion}

Accurate localization in dense urban areas is paramount in order to solve the autonomous last-mile delivery problem. Nonetheless, it still presents important challenges. We have explored the possibilities for localization in a city environment using 3D lidar data complemented with GNSS and inertial data using a delivery robot from JD. We have shown the accuracy of different approaches, assuming that a map of the operation area is given in the form of a point cloud. In addition, we have presented a strategy for situations where the map might be corrupted or the scenario might have undergone significant changes that rendered the map outdated. We have shown that 3D scan matching is the best approach for localization when properly complimented with IMU data within an unscented Kalman filter, and GNSS data. In future work, we will explore the possibilities of integrating visual data for odometry, as well as using the lidar odometry together with the inertial data within the Kalman filter in order to increase the NDT localization accuracy.

\section*{Acknowledgment}
This work has been supported by NSFC grant No. 61876039, and the Shanghai Platform for Neuromorphic and AI Chip (NeuHeilium).

\bibliographystyle{IEEEtran}
\bibliography{ref_full}

\end{document}